\documentclass[ceqn,11pt]{wlscirep}
\usepackage[utf8]{inputenc}
\usepackage[T1]{fontenc}
\usepackage[linesnumbered,ruled,vlined]{algorithm2e}
\usepackage{subcaption}
\usepackage{graphicx}
\usepackage{bm}
\usepackage{multirow}
\usepackage{booktabs}

\SetKwInput{KwInput}{Input}                
\SetKwInput{KwOutput}{Output}              
\SetKwComment{Comment}{$\triangleright$\ }{}

\title{Semi-parametric Memory Consolidation: Towards Brain-like Deep Continual Learning}

\author[2,3$\dagger$]{Geng Liu}
\author[2$\dagger$]{Fei Zhu}
\author[2,3]{Rong Feng}
\author[5]{Zhiqiang Yi}
\author[3]{Shiqi Wang}
\author[1,2,4*]{Gaofeng~Meng}
\author[1,2*]{Zhaoxiang~Zhang}
\affil[1]{State Key Laboratory of Multimodal Artificial Intelligence Systems, Institute of Automation, Chinese Academy of Sciences, Beijing, China}
\affil[2]{Centre for Artificial Intelligence and Robotics, Hong Kong Institute of Science and Innovation, Chinese Academy of Sciences, Hong Kong SAR, China.}
\affil[3]{Department of Computer Science, City University of Hong Kong, Hong Kong SAR, China.}
\affil[4]{School of Artificial Intelligence, University of Chinese Academy of Sciences, Beijing, China}
\affil[5]{The Department of Neurosurgery, Peking University First Hospital, Beijing, China}

\affil[$\dagger$]{These authors contributed equally to this work}
\affil[*]{Corresponding author: zhaoxiang.zhang@ia.ac.cn and gfmeng@nlpr.ia.ac.cn}

\begin{abstract}
Humans and most animals inherently possess a distinctive capacity to continually acquire novel experiences and accumulate worldly knowledge over time. This ability, termed continual learning, is also critical for deep neural networks (DNNs) to adapt to the dynamically evolving world in open environments. However, DNNs notoriously suffer from catastrophic forgetting of previously learned knowledge when trained on sequential tasks. In this work, inspired by the interactive human memory and learning system, we propose a novel biomimetic continual learning framework that integrates semi-parametric memory and the wake-sleep consolidation mechanism. For the first time, our method enables deep neural networks to retain high performance on novel tasks while maintaining prior knowledge in real-world challenging continual learning scenarios, e.g., class-incremental learning on ImageNet \cite {ILSVRC}. This study demonstrates that emulating biological intelligence provides a promising path to enable deep neural networks with continual learning capabilities.	
\end{abstract}

\begin{document}
\flushbottom
\maketitle

\section*{Introduction}
Humans are excellent continuous learners, demonstrating a remarkable ability to gracefully integrate new information into existing knowledge structures throughout their lifespan \cite{kudithipudi2022biological}. This ability to learn successive tasks in sequential fashion without necessarily forgetting past experience, named continual learning (CL) \cite{dohare2024loss, hadsell2020embracing}, is also a fundamental requirement for artificial intelligence systems to be deployed in the open environment. Mastering this capability is seen as a crucial step toward creating artificial agents that can be regarded as genuinely intelligent \cite{Kumaran2016}.
For instance, an autonomous vehicle initially trained on urban data will encounter diverse landscapes, e.g., forests, deserts, and countryside, and the system must adapt to new environments while retaining previously learned behaviors \cite{verwimp2023clad}. 
Unfortunately, although deep neural networks (DNNs) outperform humans in specialized tasks, such as complex games \cite{silver2017mastering} and pattern recognition \cite{zhang2020towards}, their impressive performance largely relies on stationary scenarios. When exposed to dynamic data, DNNs suffer catastrophic forgetting \cite{McCloskey1989, Goodfellow2013}, i.e., the performance on previously learned tasks declines sharply after new task acquisition.

To enable continual learning ability for DNNs, some studies \cite{Kirkpatricka2017, wang2023incorporating, zhang2023brain} mimic synaptic stabilization to mitigate catastrophic forgetting by quantifying and constraining critical parameters during the continual learning process. 
The idea of context-dependent gating \cite{heald2023computational} has also been explored \cite{masse2018alleviating, tsuda2020modeling, van2020brain} to alleviate interference between current and previous tasks. Although effective in the task-incremental learning (task-IL) scenario where task identity is provided at inference, these approaches fail in the more practical class-incremental learning (class-IL) scenario, where models must distinguish classes across tasks without task identity \cite{van2022three, van2020brain}. 
Another straightforward and effective solution is to store and replay raw data of previous tasks when learning new tasks. However, such replay is undesirable in practice due to safety or privacy concerns \cite{price2019privacy, lee2022public}, and from a neuroscience perspective, the brain is unlikely to store the raw data (e.g., pixels of an image). In practical scenarios, to prevent excessive storage space consumption as the number of tasks continues to grow, the number of exemplars stored per task must be strictly limited, thus the performance of the methods remains constrained.
Generative replay-based methods tried to overcome these limitations by training a generative model, while the quality of generated images gradually declines as the task progresses. Thus, these methods still struggle to scale to real-world class-IL challenges \cite{van2020brain}.

\begin{figure}[!t]
  \centering
  \includegraphics[width=1\textwidth]{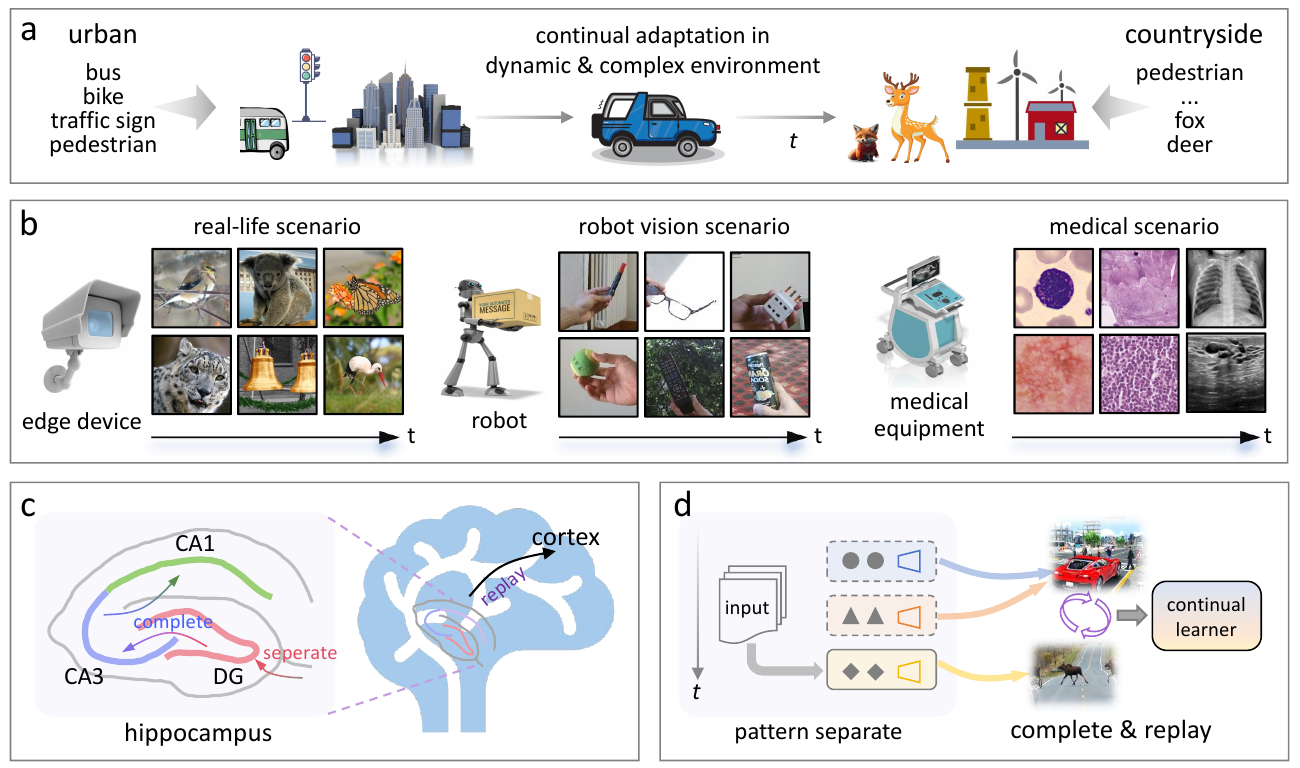}
  \vspace{-15pt}
  \caption{\textbf{Brain-inspired continual learning system.} \textbf{a}, An example of the continual learning scenario, where an autonomous car initially trained on urban data needs to adapt to the new environment with novel objects.
  \textbf{b}, Edge devices, robotic systems, and medical equipment could encounter dynamic tasks and data streams in operational environments. 
  Due to constraints in storage and computational capacity, they cannot save abundant raw samples and perform model retraining frequently to adapt to emerging tasks and data streams.
  BrainCL addresses this problem by enabling effective continuous learning under strict storage constraints, and demonstrates near-optimal performance on diverse datasets spanning real-life, robotic vision, and medical scenarios.
  \textbf{c}, In the brain, when learning new tasks, context-separated memory is formulated with pattern separation and completion in the hippocampus, i.e., the dentate gyrus (DG) is responsible for encoding new knowledge into compact episodic representations, which can be used to construct complete memory by CA3. During sleep, episodic memories are replayed and consolidated into the cortex without any external input. \textbf{d}, Inspired by the above mechanism in the brain, we propose to incorporate context-separated memory and sleep replay for a continual learner to achieve the goal of learning without forgetting.}
  \label{idea}
  \vspace{-10pt}
\end{figure}
Inspired by human continual learning, we turn to neuroscience for principles to address catastrophic forgetting. The human brain employs a complementary memory system \cite{Kumaran2016}, integrating two distinct subsystems critical for continual learning: hippocampus-based encoding of episodic memory and neocortex-based learning of structured knowledge. Here we highlight two corresponding mechanisms. The first mechanism is context-dependent memory encoding. 
The hippocampus specializes in the rapid encoding of episodic information \cite{tulving2002episodic}, forming working memory of learned experiences.
Particularly, the hippocampus is thought to encode learned experiences in a pattern-separated manner, with the environmental context represented by a separate neural space \cite{treves1992computational, alme2014place}. Therefore, it is unreasonable to view the hippocampus as a raw data-based memory buffer or single generative neural network \cite{Shin2017, rolnick2019experience}. 
The second key mechanism is the sleep-dependent consolidation cycle \cite{walker2004sleep, ji2007coordinated}. During wakefulness, the brain acquires task-specific memories, e.g., novel object recognition \cite{mallory2025time}. 
During the sleep phase, episodic memories are subsequently replayed and consolidated into the neocortex for long-term memory without any external input \cite{singh2022model}. Besides, the memory can be actively organized during non-rapid eye movement (NREM) sleep, enabling the removal of unimportant or outdated information \cite{cairney2024forgetting}. The above wake-sleep mechanism facilitates graceful continual learning ability of the brain, while been overlooked by existing continual learning works.

In this study, we explore to reduce catastrophic forgetting for DNNs by mimicking those neuroscience principles. To this end, we develop a novel semi-parametric memory framework that integrates the context-dependent memory encoding and sleep-dependent consolidation in artificial neural networks, which we refer to as brain-inspired consolidation (BrainCL).
This approach firstly encodes non-parametric entropy-sparse memory along with a parametric context-dependent pattern completion network for each task, to obtain a high-efficient memory module. 
Then it divides the training process into two functionally distinct phases: the wake phase and the sleep phase. During the wake phase, it encodes newly emerged input information to construct the working memory and leverages this memory to efficiently train the model. During the sleep phase, working memory is selectively transferred into long-term memory, and then the model undergoes consolidation through internal memory replay without external inputs.
For the first time, we show that our method can achieve nearly optimal performance on the challenging class-IL setting on large-scale benchmarks like ImageNet \cite{ILSVRC} with a strictly limited memory buffer. We also demonstrate the effectiveness of our method on robotic vision and medical datasets. Besides, our method is remarkably robust and reliable in more complex scenarios with long-tailed, noisy, and unexpected input data.

\begin{figure}[!t]
  \centering
  \includegraphics[width=\textwidth]{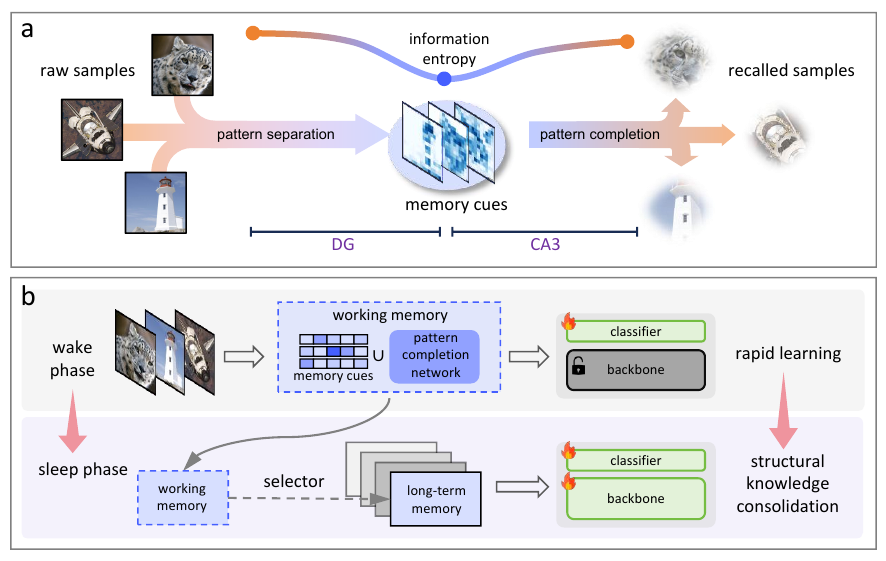}
  \vspace{-15pt}
  \caption{\textbf{Overview of our approach.} 
  \textbf{a}, The memory module first processes raw samples through pattern separation to generate corresponding memory cues with low information entropy. Samples could be recalled from the memory cues through pattern completion. This operational mechanism mirrors the complementary functions of the dentate gyrus (DG) and CA3 regions in the hippocampus.
  \textbf{b}, BrainCL contains two distinct learning phases, namely the wake phase and the sleep phase. During the wake phase, samples of the new task are memorized into semi-parametric working memory, composed of non-parametric memory cues and a parametric pattern completion network. Only the classifier is trained to acquire classification abilities for the new task.
  During the sleep phase, working memory is selectively transferred into long-term memory, based on which the entire model is finetuned without external inputs to perform structural knowledge consolidation.} 
  \label{Pipeline}
  \vspace{-5pt}
\end{figure}

\section*{Results}
Our goal is to develop a brain-inspired continual learning method to alleviate catastrophic forgetting in DNNs. We mainly focus on the practical class-IL scenario, where each task contains a group of classes, and all classes from learned tasks have to be distinguished without knowing the task identification. To this end, we propose a novel and principled approach including a semi-parametric memory module and the wake-sleep consolidation.

\subsection*{Semi-parametric Memory}
Memory is a cornerstone of biological intelligence, enabling humans and animals to integrate new knowledge without overwriting previously learned information. For continual learning of DNNs, two memory-based replay mechanisms have been explored, named data replay \cite{rebuffi2017icarl, rolnick2019experience} and generative replay \cite{Shin2017, van2020brain}. Specifically, data replay parallels the hippocampus to a memory buffer that stores raw past data and reuses it during training. However, from a neuroscience perspective, human memory is not perfect \cite{quiroga2008sparse} and the brain is unlikely to directly store raw data.
In generative replay, the hippocampus is modeled as a generative neural network, where replay is treated as a process of generating past experiences. However, there is no context information in the generative model and the weights can be easily overwritten, losing the information about previously learned tasks. Therefore, generative replay is still struggling in large-scale Class-IL with complex inputs.
Previous neuroscience research has demonstrated that the dentate gyrus (DG), CA3, and CA1 subregions of the hippocampus play pivotal roles in memory formation and neural information processing.
Among these, a key function of the DG is to mediate pattern separation, i.e., the process that transforms sensory inputs from the entorhinal cortex into distinct, sparse representations with balanced representational richness and energy efficiency \cite{benna2016computational, benna2021place}. While the CA3 subregion enables pattern completion, i.e., the process that recalls complete memories from partial or degraded representations. By linking episodic memories with context-dependent information \cite{dudai2004neurobiology}, the CA3 subregion allows for flexible and context-specific recall.

Inspired by the neuroscientific mechanisms mentioned above, as shown in Fig. \ref{Pipeline}a, we propose a semi-parametric memory (SPM) for continual learning, comprising non-parametric low-entropy memory cues and task-dependent parametric pattern completion networks. On the one hand, the non-parametric part stores arithmetic-encoded memory cues for each sample. Specifically, arithmetic coding encodes information into a binary sequence, whose length can reach the lower bound declared by Shannon's source coding theorem. The entropy of memory cues is systematically minimized to eliminate redundant information, thereby minimizing the length of the encoded binary sequence and improving the storage efficiency. On the other hand, the parametric part incorporates lightweight task-specific pattern completion networks to store task-specific contextual information and recall samples by completing corresponding memory cues.
During memory module training, we utilize a temporary auxiliary pattern separation network to synthesize memory cues directly from raw samples. 
Both pattern separation and pattern completion networks are end-to-end trained through backpropagation under the supervision of two losses. Specifically, Recall loss is computed as the mean squared error between raw samples and recalled samples, while Entropy loss quantifies the information entropy of memory cues. By jointly minimizing these two losses during training, we achieve a substantial reduction in information redundancy while retaining essential semantic information.

For every task in continual learning scenarios, the semi-parametric memory module eventually preserves two essential components: 1) a set of non-parametric low-entropy memory cues that efficiently encode sample-specific information, and 2) task-dependent parametric networks dedicated to context-aware pattern completion.
The proposed semi-parametric memory module adopts a neurologically inspired dual-component architecture that preserves both memory cues and their corresponding pattern completion networks. This design on the one hand prevents cross-task memory interference and overwriting, while on the other hand ensures high-quality memory recall in demanding Class-IL scenarios. Entropy-based redundant information elimination fosters efficient memory cues and mitigates issues related to storage demands associated with saving raw samples.

\subsection*{Wake-sleep Consolidation}
In the above, we focus on memory construction when learning a specific task. Here we explore how to build and integrate knowledge from different tasks over time. 
Studies in animals and humans have demonstrated that learning not only occurs during wakefulness but also during sleep, with the latter being crucial for memory and knowledge consolidation. Specifically, spontaneous neuronal activity reactivates learned experiences without external input, enabling the brain to form associations between old and new experiences, integrating memories into broader semantic structures \cite{singh2022model, stickgold2013sleep}. The brain's sleep replay system offers inspiration for overcoming catastrophic forgetting in artificial neural networks.

Based on the above mechanism, we propose a two-stage continual learning paradigm, as illustrated in Fig. \ref{Pipeline}b.
The first stage is termed the wake learning phase, during which the memory module is initially trained on raw inputs of the new task to establish semi-parametric episodic working memory. Once established, this working memory becomes the exclusive substrate for subsequent model learning, eliminating dependence on external inputs.
With the backbone network frozen, only the classifier of the model is subsequently trained using the working memory to enable rapid acquisition of classification capability for the new task. 
The second stage is termed the sleep consolidation phase, during which portions of the working memory are selectively transferred into long-term episodic memory at first. 
The entire model is then finetuned on both new and old task samples, which are randomly recalled from long-term episodic memories without any external inputs.
This process facilitates structured knowledge integration of the model, while consolidating model capabilities across all learned tasks.

Compared to conventional one-stage methods, the advantages of the proposed two-stage framework are clear and straightforward. During the wake learning phase, freezing the backbone network while exclusively training the classifier effectively prevents catastrophic forgetting of prior knowledge. In the sleep consolidation phase, training the complete model through simultaneous replay of memories from both new and previous tasks enables structural integration of cross-task knowledge, thereby ensuring strong model capabilities in all learned tasks.
\begin{table}[t]
    \renewcommand\arraystretch{1}
    \small
    \centering
    \caption{\textbf{BrainCL achieves state-of-the-art continual learning performance.} We compare the performance of different methods on ImageNet-100 dataset, Core50 and MedMNIST datasets using different settings. $\mathrm{MS}$ denotes the size of extra memory cost (MB), $\bar{A}$ denotes the average top-1 accuracy, $A_{B}$ denotes the last top-1 accuracy. The best results are highlighted in \textbf{bold}.}
    \label{Table1}
    \begin{tabular}{l|ccccccccc}
    \toprule
    \multirow{2}{*}{Method} 
            & \multicolumn{3}{c}{ImageNet-100}  & \multicolumn{3}{c}{Core50} & \multicolumn{3}{c}{MedMNIST}\\
            & $\mathrm{MS} \downarrow$ & $\mathcal{\bar{A}} \uparrow$ & $\mathcal{A_{B}} \uparrow$ & $\mathrm{MS} \downarrow$ & $\mathcal{\bar{A}} \uparrow$ & $\mathcal{A_{B}} \uparrow$ & $\mathrm{MS} \downarrow$ & $\mathcal{\bar{A}} \uparrow$ & $\mathcal{A_{B}} \uparrow$\\
    \midrule
    Joint   &   -   & 84.2  & 82.9  &      -   & 86.1  & 86.8  &   -   & 86.3  & 81.8 \\
    \midrule
    Finetune&   -   & 26.3  & 9.2   &      -   & 26.7  & 9.7   &   -   & 28.0  & 9.0 \\
    EWC     &   -   & 28.3  & 10.9  &      -   & 28.3  & 10.0   &   -   & 28.3  & 8.4 \\
    Replay  &  287  & 49.4  & 39.6  &    72  & 51.7  & 47.8   &  115  & 64.4  & 54.7 \\
    iCaRL   &  287  & 61.4  & 48.2   &    72  & 63.3  & 53.3   & 115   & 69.8  & 60.3 \\
    DER  &  672 & 71.3  & 63.5   &    457  & 69.6  & 60.1  &  500  & 75.9  & 66.7 \\
    BrainCL & \textbf{280} & \textbf{78.7} & \textbf{74.2} & \textbf{70} & \textbf{79.0} & \textbf{78.4} & \textbf{97} & \textbf{83.3} & \textbf{76.2} \\  
    \bottomrule
    \end{tabular}
\end{table}

\subsection*{Practical Continual Learning Evaluation}
Previous work typically evaluated on small and easy datasets like MNIST \cite{deng2012mnist} and CIFAR \cite{krizhevsky2009learning}. In this study, we demonstrate the superiority of BrainCL in the challenging class-incremental learning scenario. To this end, we test our method on the standard large-scale, natural image dataset with complex inputs like ImageNet \cite{ILSVRC}. We also report the performance on the robot view dataset and medical dataset. Furthermore, considering that real-world applications often involve environments that can change easily, we propose to verify the robustness of the continual learner to covariate-shifted and long-tailed distributions. Besides, unknown classes would emerge at any time, and it is important that the model can reject them safely before learning them. We hope our study could also motivate further research on studying important aspects of continual learning such as robustness, reliability and fairness.
\begin{figure}[!t]
  \centering
  \includegraphics[width=0.97\textwidth]{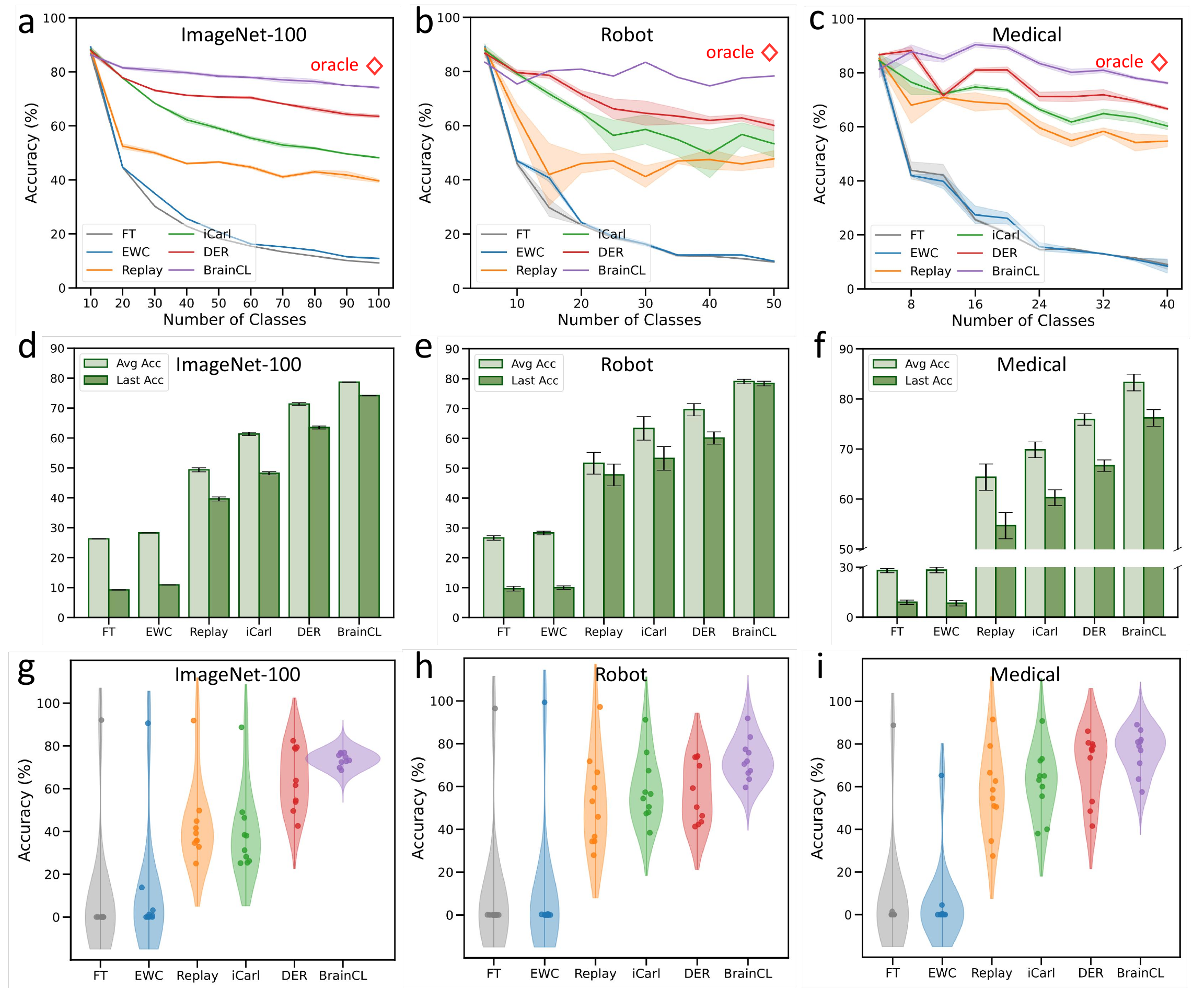}
  \caption{\textbf{BrainCL enhances the performance of continual learning on large-scale and realistic datasets.} Test accuracy for incrementally learned classes on \textbf{a} natural image dataset ImageNet-100, \textbf{b} robot vision dataset CoRe50, and \textbf{c} medical dataset MedMNIST, presented as means over three random seeds with shaded areas indicating $±$SEM. Average and last accuracy of class-incremental learning on \textbf{d} ImageNet-100, \textbf{e} CoRe50 and \textbf{f} MedMNIST datasets. \textbf{d} The distribution of test accuracy scores for all tasks in the \textbf{g} ImageNet-100, \textbf{h} CoRe50 and \textbf{i} MedMNIST, with width representing probability density and overlaid scatter points indicating individual data points. }
  \vspace{-10pt}
  \label{fig3}
\end{figure}

\subsection*{Large-scale Continual Learning with Complex Inputs} 
Existing brain-inspired continual learning studies \cite{Kirkpatricka2017, wang2023incorporating, zhang2023brain} focus on the easy task-incremental learning scenario \cite{van2022three} and typically evaluated on small datasets with low resolution and limited classes \cite{deng2012mnist, krizhevsky2009learning}. To explore the potential of the brain-inspired method in realistic scenarios, we evaluate our proposed BrainCL approach on the large-scale, challenging Class-IL scenario with complex inputs. In this scenario, the model continually learns multiple classes from different tasks and must recognize all previously learned classes ultimately. Specifically, we conduct experiments on ImageNet \cite{ILSVRC}, which is a challenging and large-scale visual classification dataset containing 1000 classes with high-resolution (224 $\times$ 224) natural images. To demonstrate the effectiveness of our method on more applications, we also evaluate it on a robot view dataset CoRe50 \cite{lomonaco2017core50} and medical dataset MedMNIST \cite{yang2023medmnist}. The popular deep neural network named ResNet-18 \cite{he2016deep} is trained from scratch in an end-to-end manner on those benchmarks (see Methods and Supplementary Information for details). We denote the test accuracy after the $i$-th task as $\mathcal{A}_i$, then the average accuracy $\bar{\mathcal{A}}=\frac{1}{B}\sum_{i=1}^{B}\mathcal{A}_i$ represents model’s
average performance during the continual learning process, and the last accuracy $\mathcal{A_{B}}$ denotes the performance on all learned classes after the last learning stage. 
\begin{figure}[!t]
  \centering
  \includegraphics[width=0.97\textwidth]{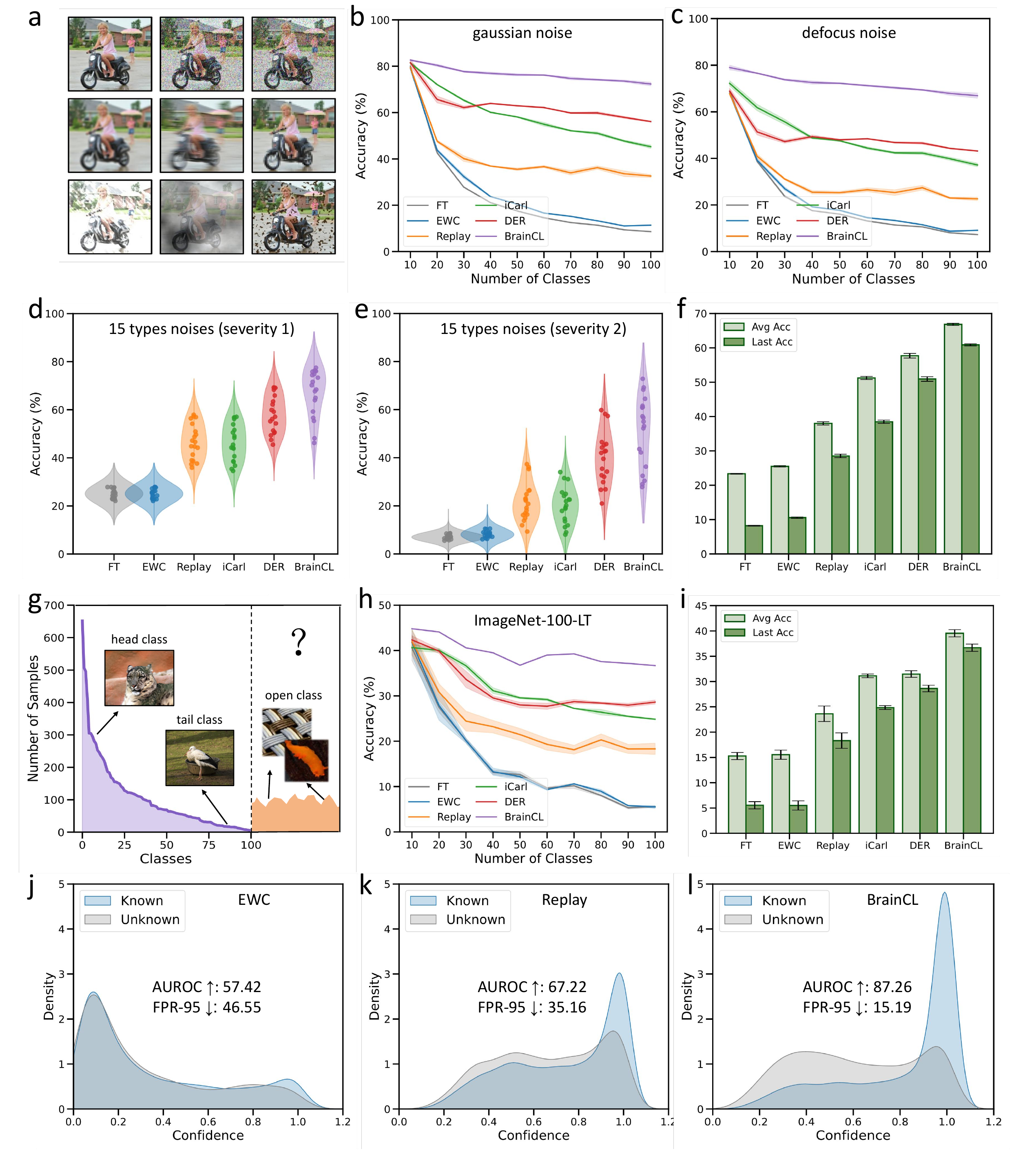}
  \vspace{-5pt}
  \caption{\textbf{BrainCL enhances the robustness and reliability of continual learning on more practical scenarios.} \textbf{a} Examples from the corrupted dataset, which consists of 15 types of corruptions from noise, blur, weather, and digital categories. Test accuracy for incrementally learned classes on corrupted ImageNet-C dataset with \textbf{b} Gaussian noise and \textbf{c} Frost noise. \textbf{d-e} Test accuracy of 15 types of corruptions with two different levels of severity. \textbf{f} Average and last accuracy of Class-IL on ImageNet-C. \textbf{g} Distribution of long-tailed classes unknown classes that could be encountered in the open world. \textbf{h} Test accuracy for incrementally learned classes on ImageNet-LT. \textbf{i} Average and last accuracy of Class-IL on ImageNet-LT. Confidence distributions of known and unknown classes after Class-IL with \textbf{j} Replay and \textbf{k} BrainCL approaches. \textbf{l} Comparison of reliability metric, e.g., AUROC among different methods.}
  \vspace{-10pt}
  \label{fig4}
\end{figure}

The comparative results of our method with other representative approaches are shown in Fig. \ref{fig3} and Tab. \ref{Table1}. We can observe that plasticity-inspired methods like EWC \cite{Kirkpatricka2017} dramatically failed in the Class-IL scenario, though they are effective in the Task-IL scenario. This observation has also been concluded by other studies \cite{van2020brain, van2022three}. We also find that generative replay performs poorly on real large-scale datasets mainly due to two reasons: on the one hand, generative high-resolution image is challenging; on the other hand, the memory of old tasks stored in the generative model would overwritten after learning new tasks. That is, the generative model itself also suffers from catastrophic forgetting. Replay \cite{rolnick2019experience}, iCarl \cite{rebuffi2017icarl} and DER \cite{yan2021dynamically}, which are representative data replay-based methods. Particularly, DER is the state-of-the-art method that combines data replay with continual network expansion. However, there still exists an obvious performance gap between them and the joint training (which is viewed as the upper bound for continua learning). Differently, BrainCL achieves consistently the strongest performance, nearly approaching that of joint training, as shown in Fig. \ref{fig3}a. Class-IL results (Fig. \ref{fig3}b-c) on robot view dataset CoRe50 \cite{lomonaco2017core50} and medical dataset MedMNIST \cite{yang2023medmnist} also valid the remarkable superiority of BrainCL, indicating that our method is applicably in a wide application involving nature images based system (e.g., autonomous system), robot and medical diagnosis. The results of average and last accuracy in Fig. \ref{fig3}d-f further confirm the above conclusion. Moreover, we visualize the accuracy of the final task and the average accuracy of the previous tasks, which can indicate the plasticity and stability of the continual learner. As shown in Fig. \ref{fig3}g-i, existing methods experience varying degrees of imbalance across different tasks, while BrainCL successfully maintains the balance between stability and plasticity. Finally, we explore how our method performs when learning very long sequent tasks.

\subsection*{Robustness and Reliability in More Practical Scenarios} 
The experiments mentioned above are conducted on the datasets with clean samples. However, the input can suffer from distribution shifts for a continual learner deployed in the open world. On the one hand, a continual learner may encounter covariate shifts such as conditions with snowy night \cite{sakaridis2021acdc} or corrupted inputs resulting from camera noise and sensor degradation \cite{hendrycks2018benchmarking}, as shown in Fig. \ref{fig4}a. Naturally, we expect the model to suffer less performance deterioration when deployed in those scenarios. On the other hand, unknown categories with semantic shifts may also emerge (Fig.~\ref{fig4}g), and we expect the model can reliably handle them before continually learning them, e.g., reject to make a prediction and trigger warning or hand over the input to human experts. Moreover, classes in the real world often follow a long-tailed distribution (Fig. \ref{fig4}g), e.g., the normal samples are typically more than the disease samples in a disease diagnosis system. Therefore, we propose to verify the robustness and reliability of a continual learner. 

We validate the robustness of BrainCL when facing covariate-shifted dataset ImageNet-C \cite{hendrycks2018benchmarking}, long-tailed dataset ImageNet-LT, and semantic-shifted dataset ImageNet-O \cite{hendrycks2021natural}. 
First, we test the model's robustness to distribution shifts such as brightness, snow, and fog. At each continual stage, the model is trained on a clean train set but tested on a corrupted test set. For example, a continual learning model is trained on data collected in sunny, while being deployed on rainy days.
Fig. \ref{fig4}b-c report comparative results on two types of noises (Gaussian and Frost noise), in which the performance of all methods suffer from degradation compared with that in Fig. \ref{fig3}a. However, BrainCL is significantly more robust than other baselines, which can be further confirmed by the results on all 15 types of noises in Fig. \ref{fig4}d-f. Next, we perform continual learning on the long-tailed dataset ImageNet-100-LT, where the number of samples is heavily imbalanced across classes during training (Fig. \ref{fig4}g), e.g, a majority class has 1000 samples while a tail class only has 5 samples. The results in Fig. \ref{fig4}h-i demonstrate that BrainCL is more robust than other approaches when facing long-tailed data. Finally, we consider a more general setting where wild samples from unseen classes could emerge during inference time. An example is that for a model trained with natural images, digital samples or images with pure noise that belong to neither old classes nor current classes could be inputted. Deep models should ``know what they do not know'' based on the models' confidence in given inputs and reject to make any prediction on them for safety. We compare the reliability of BrainCL with other baselines and the results are summarized in Fig. \ref{fig4}j-l, verifying that BrainCL performs the best. Specifically, the visualization of confidence (i.e., the maximum softmax probability) distribution in Fig. \ref{fig4}j and Fig. \ref{fig4}k demonstrate that BrainCL performs much better when distinguishing unknown from known classes, which can also be verified via the reliability metric like AUROC in Fig. \ref{fig4}l. 
In conclusion, our method is remarkably robust and reliable in more practical scenarios involving covariate and semantic distribution shifts.

\begin{figure}[!t]
  \centering
  \includegraphics[width=0.95\textwidth]{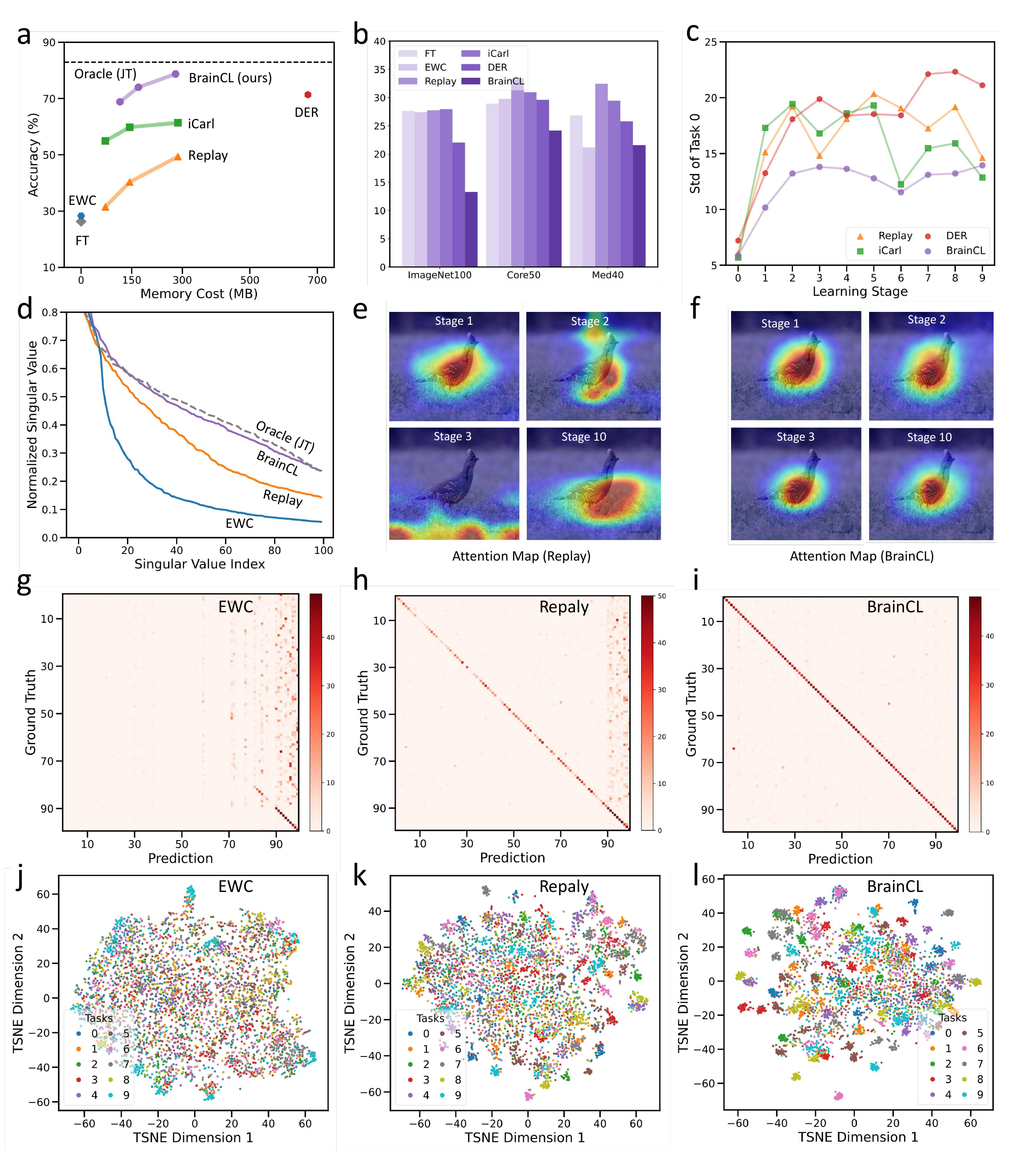}
  \vspace{-15pt}
  \caption{\textbf{Understanding BrainCL with further analysis.} \textbf{a} Memory comparison: BrainCL achieves strong performance with limited memory cost, while DER has a large memory due to continual backbone expansion. \textbf{b} Class-wise fairness across all learned classes at the end of CL. \textbf{c} We reveal that continual learning can cause the disparate impact of different classes in the same task (e.g., task 1), and BrainCL leads to less disparate impact. \textbf{d} EWC and replay lead to feature space collapse, while our BrainCL maintains the feature space as well as the oracle, i.e., joint training. Comparison of attention regions during continual learning with \textbf{e} EWC and \textbf{f} BrainCL. Confusion matrix across the 100 classes in the ImageNet-100 dataset with \textbf{g} EWC \textbf{h} Replay and \textbf{i} BrainCL. TSNE visualization of the continual models trained with \textbf{j} EWC \textbf{k} Replay and \textbf{l} BrainCL.}
  \vspace{-10pt}
  \label{fig5}
\end{figure}
\subsection*{Understanding BrainCL with Further Analysis}
To further understand BrainCL, we provide analysis in terms of memory cost, fairness, feature space and other informative visualizations during continual learning. First, we compare the memory cost of different methods in \ref{fig5}a. Take experiments on ImageNet-100 dataset as an example, we set the number of saved samples of replay-based methods to 20 per class and 2000 in total, thus the total memory cost is 329 MB and the extra memory cost (memory cost besides the backbone ResNet-18) is 287 MB. The extra memory cost of our method is 280 MB which is similar to Replay and iCarl, while our method performs significantly better than them. Particularly, DER has a much higher memory cost since it continually expends the whole backbone at each stage, while still performing worse than ours. Besides, as also demonstrated in \ref{fig5}a, the performance of BrainCL is much more robust than others when reducing the memory of all methods. Second, we reveal that existing continual learning methods tend to deteriorate the unfairness issue. On the one hand, class-wise accuracy of all learned classes varies at the end of CL (\ref{fig5}b), which is a form of unfairness; on the other hand, there exists a larger disparity of accuracy among different classes in the same task (\ref{fig5}c), indicating the disparate impact of continual learning methods. For example, a model has similar accuracy (80\%) on \textit{cat} and \textit{dog} in the first task. After learning the other four tasks with state-of-the-art CL algorithm, the model has 70\% accuracy on \textit{cat} but only 5\% accuracy on \textit{dog}. As shown in Fig. \ref{fig5}b-c, our proposed BrainCL effectively alleviates the unfair and disparate impact issue. 

We investigate the deep feature space learned with different CL methods via spectral decomposition. Specifically, given a continual model learned on all classes and the corresponding test set, we get the deep features in the final deep space. Using eigenvalue decomposition, the feature space of the model can be characterized by the distribution of eigenvalue and eigenvector, and the number of directions with significant eigenvalues indicates the diversity or capacity of the feature space. As shown in Fig. \ref{fig5}d, EWC suffers significantly from feature space collapse, where only a limited directions remains. While Replay effectively alleviates the feature space collapse, there still remains a gap compared with the oracle model (i.e., jointly trained model). BrainCL performs similar with the oracle model, avoiding the loss of feature diversity or capacity compared with joint training. Fig. \ref{fig5}e-f show the  attention regions in CL, and we find that existing methods lead to attention degrade, while with BrainCL the attention regions do not vary across continual learning steps. Fig. \ref{fig5}e-i display the confusion matrix, we find that EWC and Replay tend to classify the samples into new classes, resulting in strong confusions on the last task. BrainCL is capable to remove most of the bias
and achieves better overall performance. The TSNE \cite{van2008visualizing} visualization in Fig. \ref{fig5}j-l further demonstrate the superiority of BrainCL in maintaining the discriminability among continually learned classes.

\section*{Discussion}
Continual learning is a critical ability for primates to accommodate and adapt to dynamic real-world environments. However, this remains a significant challenge for modern artificial intelligence. Current brain-inspired continual learning approaches predominantly focus on synaptic plasticity mechanisms \cite{Kirkpatricka2017, wang2023incorporating, zhang2023brain}, where importance metrics are designed to quantify parameter relevance for current tasks, followed by constrained parameter updates during new task acquisition. While theoretically plausible, these methods exhibit notable limitations in class-incremental learning scenarios, suffering from catastrophic performance degradation. Modern neuroimaging studies further demonstrate that prefrontal cortical areas mediate task-specific pattern separation through neurogenesis and dendritic spine turnover \cite{frankland2019neurobiological}, enabling biological systems to compartmentalize new knowledge without overwriting existing representations. Current synaptic importance-based algorithms lack such structural plasticity mechanisms, failing to replicate the brain's capacity for dynamic neural resource allocation. Emerging insights into memory reconsolidation further challenge current synaptic plasticity-based continual learning paradigms. Neural reactivation studies reveal that memory retrieval triggers transient plasticity windows allowing memory updating through neuromodulatory signals \cite{nader2000fear}, a process fundamentally different from static parameter regularization. Moreover, the discovery of engram cells exhibiting experience-specific connectivity patterns \cite{josselyn2020memory} suggests biological memories maintain modular independence through sparse neural recruitment, whereas synaptic plasticity-based methods typically distribute task information diffusely across shared parameters. This architectural divergence may underlie the poor scalability of existing brain-inspired models in Class-IL with complex inputs and more practical scenarios.

Alternative approaches inspired by the dual-memory system hypothesis attempt to reconcile rapid learning with long-term retention through complementary systems. Nevertheless, most methods \cite{rebuffi2017icarl, rolnick2019experience, yan2021dynamically} rely on direct storage of raw training exemplars, raising substantial concerns regarding memory efficiency and diversity. Notably, no neurobiological evidence confirms that mammalian brains store exact sensory replicas, e.g., the brain unlikely to directly store all pixels of an image \cite{van2020brain}; instead, recent advances in systems neuroscience reveal that biological memory consolidation involves distinct hippocampal-neocortical interactions where novel experiences initially form sparse, localized representations before gradual integration into cortical networks through offline replay mechanisms \cite{kumaran2016learning, tonegawa2018role}. This mismatch highlights a paradoxical divergence from biological plausibility even within supposedly brain-derived frameworks. Addressing these limitations may require novel architectures incorporating dynamic neural resource allocation, biologically plausible reactivation mechanisms, and multi-timescale consolidation processes informed by recent breakthroughs in memory engram manipulation and in vivo neural circuit analysis.

In this work, we develop a novel brain-inspired continual learning framework by incorporating principles drawn from human brain memory systems. One of the key contributions of our approach is the design of episodic memory. Drawing from recent discoveries in engram cell biology \cite{josselyn2020memory}, we generate entropy-sparse representations through a dentate gyrus-inspired pattern separation network, effectively minimizing interference during incremental task acquisition.
Unlike generative replay that can suffer from loss of contextual detail and task-specific information during continual learning, our system preserves task-specific memory traces via low-entropy memory cues and lightweight pattern completion network, mimicking the brain's energy-efficient encoding of episodic details \cite{richards2017persistence}. This is very important for mitigating the forgetting in the challenging large-scale Class-IL, in which replay is a key requirement. 
Additionally, the incorporation of a wake-sleep consolidation mechanism, inspired by human sleep processes, addresses the challenge of integrating new and old knowledge. During sleep, the hippocampus replays learned experiences, facilitating the formation of associations between new and prior knowledge, which is vital for continuous knowledge integration without interference. 
Our model replicates this by replaying memorized samples during a sleep phase, allowing cross-task knowledge integration in a way that mirrors biological memory consolidation.
This phase is crucial for mitigating catastrophic forgetting by stabilizing and transferring learned knowledge across tasks. These neurocomputational design collectively address the critical limitations of existing brain-inspired approaches while maintaining compatibility with modern deep learning architectures.

Experimentally, we show that the Class-IL performance could approach that of joint training on large-scale ImageNet dataset for the first time, which is a clear demonstration that mimicking biological memory consolidation is promising to equip continual learning ability for artificial neural networks. Beyond conventional accuracy metrics, we rigorously evaluate robustness under realistic operational conditions, including covariate shifts, long-tailed data distributions, and anomalous inputs—scenarios routinely encountered in dynamic real-world environments but often overlooked in existing studies. Crucially, our framework exhibits consistent robustness and reliability across these challenging settings (Fig. \ref{fig4}), outperforming existing methods that suffer from severe performance degradation under non-ideal conditions. Notably, our architecture maintains equitable knowledge retention across tasks of varying complexity, which is particularly vital for real-world applications requiring consistent performance in evolving environments.
The broader continual learning community has predominantly focused on simplified evaluations using clean and balanced datasets. Real-world deployment scenarios demand robustness against sensor noise, reliability in safety-critical domains. We emphasize the necessity of redefining evaluation protocols of continual learning systems for real-world applications. In conclusion, this study highlights the promise of leveraging biological principles of memory consolidation to address the limitations of current deep learning models in continual learning tasks, paving the way for more adaptable and intelligent artificial systems.

Contemporary continual learning studies critically overlook fundamental limitations inherent to supervised learning paradigms. Most approaches employ standard cross-entropy loss with one-hot labels when leaning each task, which is neurobiologically incongruent with how biological systems integrate knowledge. This mismatch manifests in two key issues: 1) overfitting amplification, where rigid one-hot supervision forces excessive specialization to current task features, exacerbating interference with consolidated knowledge; and 2) generalization deficiency, as maximized likelihood estimation under fixed labels discourages exploratory representation learning essential for lifelong adaptation. Actually, many studies suggest that brain can learn to predict future rewards from past experience, and use reward information for learning \cite{kasdin2025natural, schultz2000multiple}. Therefore, reward-based updates may better emulate biological learning mechanisms, and reframing supervised tasks as a reinforcement learning paradigm, where models receive scalar rewards rather than categorical targets, might be beneficial for forgetting mitigation in continual learning. Recent evidence in large models demonstrates the potential of reinforcement learning in replacing classical supervised learning. Furthermore, there is no direct evidence that the brain uses a backprop-like algorithm for learning \cite{lillicrap2020backpropagation}, which raises question about backpropagation's suitability for continual scenarios where precise gradient calculations conflict with dynamic stability-plasticity balance. 
In addition, it has been shown that biological cognition system typically employs dual-process mechanisms \cite{kahneman2011thinking}, i.e., a fast system quickly performs intuition-driven processing for familiar inputs, and a slow system engages in eliberative reasoning engaging memory retrieval and algorithmic processing. While current continual learning systems lack this adaptive specialization, applying uniform computation regardless of input complexity. Neuroimaging reveals that biological brains dynamically allocate resources, engaging hippocampal-prefrontal circuits for ambiguous stimuli while relying on cortical shortcuts for routine decisions \cite{barron2020neuronal}. Implementing such dual mechanisms could enable continual learners to achieve natural continual adaptation through hierarchical resource allocation. 
These above neurocognitive perspectives highlights critical gaps in current approaches.
Bridging these gaps requires rethinking both learning objectives and architectural principles, potentially unlocking more robust and biologically aligned continual learning frameworks.

\section*{Methods}
\textbf{Problem Formulation.}
Class-incremental learning aims to learn a stream of data continually with new classes while maintaining the discrimination ability on old classes.
Assume there are a sequence of $\mathcal{B}$ training tasks, the training set at task $t \in \mathcal{B}$ is defined as $\mathcal{D}^t=\{\bm{x}^t_i, \bm{y}^t_i\}^{n_t}_{i=1}$, in which $\bm{x}^t_i$ is the sample and $\bm{y}^t_i\in \mathcal{C}^t$ is the corresponding label, $\mathcal{C}^t$ is the class set of task $t$. Note that the class sets of different tasks are disjoint, that is $\mathcal{C}^i \cap \mathcal{C}^j=\emptyset$ for $i \neq j$. 
The model is evaluated on all of the classes ever seen $\bigcup_{i=1}^{t}\mathcal{C}^i$ after incremental learning step $t$.

\subsection*{Memory Module}
\textbf{Architecture Design.}
Pattern separation network and pattern completion network are symmetrical 4-layer convolutional neural network and deconvolutional neural network respectively. Both networks utilize ReLU activation. 
Kernel size of convolutional layers and deconvolutional layers is set to 5, the stride is set to 2, resulting the downsampling rate of each convolutional layer and the upsampling rate of each deconvolutional layer to 2. Therefore, the total downsampling rate of pattern separation network and the upsampling rate of pattern completion network are both 16.
The channels for each layer in pattern separation network are (128, 128, 128, 192) respectively, and the number of channels for each layer in pattern completion network are (128, 128, 128, 3). Dimensions of raw memory cues generated by pattern separation network depend on the size of input samples. 

\vspace{5pt}
\noindent\textbf{Loss Functions.}
We use $N_S^t$ and $N_C^t$ to denote the pattern separation network and pattern completion network of task $t$, respectively.
Specifically, for any sample $(\bm{x},\bm{y})\in\mathcal{D}^t$, the raw memory cue is $\bm{z}=N_S^t(\bm{x})$. Inspired by Balle \textit{et al.} \cite{balle2017end}, we use a uniform quantization function $Q(\cdot)$ with step size $1$ to quantize $\bm{z}$, to get the discrete-valued memory cue $\bm{\hat{z}}=Q(\bm{z})$ which is much more efficient.
Then $\bm{\hat{z}}$ could be arithmetic encoded into a binary sequence $\bm{v}$ to further reduce its number of bits $\bm{v}=A_E(\bm{\hat{z}})$, where $A_E$ and $A_D$ refers to arithmetic encoding and decoding algorithm respectively. 
When we arithmetic decode the binary sequence into quantized memory cue $\bm{\hat{z}}=A_D(\bm{v})$, the sample could be recalled through the pattern completion network: $\bm{\hat{x}}=N_C^t(\bm{\hat{z}})$. 
We note the shape of $\bm{z}$ as C$\times$H$\times$W and use $z_i$ to represent the element of $\bm{z}$, such that $\bm{z}=\{z_i\}_{i=1}^{C\times H\times W}$.
Since the quantization function is not differentiable, the model can't be updated by back-propagation directly. Because the influence of quantization to any value $z_i$ (as known as quantization noise) could be considered as adding a pseudo quantization noise $u\in[-0.5,0.5]$ to $z_i$, we could utilize $\tilde{z}_i\sim\mathcal{U}(z_i-0.5,z_i+0.5)$ to replace $\hat{z_i}$ during training, $\mathcal{U}(a,b)$ is the uniform distribution on the interval $(a,b)$. Following previous works \cite{balle2017end,widrow1996statistical,defossezdifferentiable}, we utilize the reparameterization technique to further rewrite the expression of $\tilde{z}_i$ as $\tilde{z}_i=z_i+\mathcal{U}(-0.5,0.5)$ to separate the random sampling process from the computation route, thus the computation process is derivable and $\tilde{z}_i$ could be optimized by gradient decent. 

The first objective of the memory module is to minimize the differences between the memorized samples and the realistic samples.
we utilize the Mean Squared Error (MSE) to measure the difference as the Recall loss:
\begin{equation}
\begin{aligned}
    Loss_R&=MSE(\bm{\tilde{x}},\bm{x})=MSE(N_C^t(\bm{\tilde{z}}),\bm{x})=MSE(N_C^t(\bm{z}+\bm{u}),\bm{x})=MSE(N_C^t(N_S^t(\bm{x})+\bm{u}),\bm{x}).
\end{aligned}
\end{equation}
Then, we train an entropy model to estimate the possibilities of different $\hat{z}_i$, $\psi$ is parameters of the entropy model. In detail, the entropy model utilizes piecewise linear functions to approximate the CDF (Cumulative Distribution Function) of different ${z}_i$. Thus the possibility of $\hat{z}_i$ could be estimated as:
\begin{equation}
\begin{aligned}
   p_{\psi}(\hat{z}_i)\approx p_{\psi}(\tilde{z}_i)=\int_{z_i-0.5}^{z_i+0.5}p_{\psi}(z)dz=CDF_{\psi}(z_i+0.5)-CDF_{\psi}(z_i-0.5).
\end{aligned}
\end{equation}
We therefore employ the Entropy loss function below to minimize the entropy of $\tilde{z}_i$:
\begin{equation}
\begin{aligned}
   Loss_E&=-\sum_{i}\log p_{\psi}(\tilde{z}_i)=-\sum_{i}\log(p_{\psi}(z_i+u)).
\end{aligned}
\end{equation}
This optimization indirectly reduces the entropy of the quantized memory cue $\hat{z}_i$, consequently minimizing the length of the arithmetic encoded binary sequence.
Finally, we use $Loss_R$ to supervise the pattern completion network and use $Loss_{C}= Loss_R+\lambda Loss_E$ to supervise the pattern separation network. $\lambda$ is the hyper-parameter to adjust the weight of two loss functions, thereby adjusting the quality of the memorized samples as well as the memory cost.

\vspace{5pt}
\noindent\textbf{Training Details.}
For every task, memory module is trained for 100 epochs using Adam optimizers \cite{Adam}, the batch size is set to 64. For pattern separation and completion networks, the learning rate is set to 1e-4, $\lambda$ in $Loss_{C}$ is set to 1e-2. For entropy models, the learning rate is set to 1e-3.
After memory module training, we freeze both the pattern separation and completion networks while employing the pattern separation network to generate memory cues from raw samples. Upon completion of this process, the separation network is decommissioned, leaving the memory module with solely non-parametric memory cues and the parametric completion network for subsequent operations.
Note that the resolutions of the samples in the ImageNet dataset are not uniform and there are significant differences. 
Therefore, when constructing the working memory, we did not adopt a fixed resolution. 
Instead, we dynamically adjusted the resolutions of the samples according to the resolution of the original samples. Specifically, for samples with a maximum resolution below 224, we directly generate memory cues using the original resolutions. For samples with a maximum resolution above 224, we kept their length-width ratios unchanged and normalized their maximum resolutions to 224 to generate memory cues.

\subsection*{Wake Phase}
In the wake phase, the memory module is initially trained on raw samples of the current task to construct the working memory, and then the classifier of the model will be trained rapidly based on the working memory. 
The classification process of the model could be written as: $\bm{\hat{y}}=Softmax(FC(F(\bm{\hat{x}})))$, in which $\bm{\hat{x}}$ is the recalled sample, $F(\cdot)$ is the backbone network (ResNet-18 in this paper), $FC(\cdot)$ is the fully-connected classifier, $\bm{\hat{y}}$ is the classification result. 
We use a simple cross-entropy loss to supervise the classifier:
\begin{equation}
   Loss_{CE}=-\sum_{j=0}^{C^t}y_{j}\log \hat{y}_{j},
\label{equation:CELoss}
\end{equation}
where $C^t$ is the number of learned classes at task $t$, $y_{j}=1$ if $x$ belongs to class $j$, other wise $y_{j}=0$. 
We use SGD optimizer with an initial learning rate of 0.1 to train the classifier for 10 epochs with a momentum of 0.9 and a weight decay of 2e-4.

\subsection*{Sleep Phase}
At the beginning of the sleep phase, the representative samples in the working memory are selected and transferred into long-term memory, after which the working memory is cleared out. The select algorithm is based on the feature of samples: $\bm{f}=F(\bm{\hat{x}})$. In detail, we calculate the mean feature of each class:
\begin{equation}
   \bm{\bar{f_c}}= \mathbb{E}_{(\bm{x},\bm{y})\in M_c^t}[F(\bm{\hat{x}})], 
\end{equation}
where $M_c$ is the set of memorized samples belonging to class $c$. 
Then we calculate the distance between the feature of sample $\bm{x}$ and the mean feature of the class $c$:
\begin{equation}
   d=MSE(\bm{f}, \bm{\bar{f_c}}).       
\end{equation}
We consider samples with small   $d_i^c$ representative samples and will be transferred into the long-term memory together with the pattern completion network.
Finally, all of the memorized samples in long-term memory will be utilized to train the entire model using the cross-entropy loss in Eq.~\ref{equation:CELoss}, to integrate structural knowledge and consolidate model capabilities across all learned tasks.
We use SGD optimizer to train the entire model for 60 epochs with a momentum of 0.9 and the weight decay is set to 2e-4.
The initial learning rate is set to 1e-1, we employ a multi-step learning rate scheduler to perform learning rate decay with a factor of 1e-1 at the 20th and 40th training epochs.

\subsection*{Experimental Setup}
\vspace{5pt}\textbf{Baseline methods.} 
\textbf{Finetune} is the typical baseline in class-IL, it trains the model directly on the data of the current task without using any additional tricks or techniques. Serious catastrophic forgetting often occurs when training the model in this way.
\textbf{EWC} \cite{Kirkpatricka2017} is the first parameter-regularization-based method, which maintains an importance matrix with the same scale of the network to consolidate important parameters during training, therefore reducing the forgetting of knowledge. It utilizes the Fisher information matrix to calculate the importance of every model parameter on previous tasks.
Methods mentioned above are non-exemplar methods that require no extra memory. However, a lot of works in recent years demonstrate that saving extra exemplars or parameters in class-IL could improve the performance of the model greatly.
\textbf{Replay}\cite{rolnick2019experience} is a strong baseline in class-IL with extra memory budget. For every task in class-IL, it randomly saves a fixed number of exemplars for every class to be replayed in subsequent tasks, to maintain the knowledge of the model on current task, thus the catastrophic forgetting could be mitigated greatly. 
\textbf{iCarl}\cite{rebuffi2017icarl} proposes herding-based prioritized exemplar selection method, which chooses exemplars to make the average feature vector over all exemplars best approximate the average feature vector over all training examples. Furthermore, iCarl adopts the nearest-mean-of-exemplars rule to conduct classification and utilizes knowledge distillation to improve the representation learning.
Besides the baselines and early methods introduced above, we also compared our method with some recent state-of-the-art methods as follows.
Finally, \textbf{Oracle} is a simple joint training method, which uses all the data of the current task and previous tasks for joint training in each task. The model trained in such manner could achieve the best performance. Therefore, its results on different datasets can serve as the upper bound of the performance of the current model structure on that dataset.

\vspace{5pt}
\noindent\textbf{Implementation Details.}
For all baseline methods, we employ a standard ResNet-18 architecture \cite{he2016deep} as the backbone network with randomly initialized weights. Data preprocessing includes standard augmentation techniques such as random horizontal flipping and random cropping during training. 
Models are trained using the SGD optimizers with a momentum of 0.9 and weight decay of 2e-4 at an initial learning rate of 1e-3. Training parameters are consistent across all datasets: the recognition model undergoes 200 epochs for the first task and 70 epochs for subsequent tasks with a batch size of 128, using a multi-step learning rate scheduler.

\vspace{5pt}\noindent\textbf{Large-scale ImageNet Dataset.} 
ImageNet-100 is a notable subset of the extensive ImageNet-1K dataset. It encompasses 100 distinct classes of images, ranging from animals like dogs, cats, and birds to household objects like cups and chairs, as well as vehicles and numerous other real-world entities. Each class is represented by multiple images to support the training and evaluation of machine learning models. Primarily, it serves the purpose of training and testing computer vision algorithms and deep learning models. With its images sourced from diverse real-life scenarios, featuring different lighting, angles, backgrounds, and scales, it offers a challenging yet practical dataset for models to learn from and generalize effectively. Moreover, it has been meticulously curated and labeled to guarantee the precision of class information for each image, which is vital for supervised learning tasks. The ImageNet-100 dataset contains 128,856 samples for training and 5,000 samples for testing. 
We split 100 classes into 10 tasks and each task contains 10 classes. 
For replay-based baselines, we set the number of saved samples to 20 per class and 2,000 in total, thus the extra memory cost (memory cost besides the backbone ResNet-18) of Replay and iCarl are 287 MB. Besides, DER saves the parameters of the backbone at each task, so its extra memory cost is 672 MB. In BrainCL, the memory cost is 280 MB.

\vspace{5pt}
\noindent\textbf{Robotic Vision Dataset.}
The Core50\cite{lomonaco2017core50} dataset is a robotic vision dataset created specifically for Continuous Object Recognition. It encompasses 50 household objects that are grouped into 10 distinct categories, such as plug adapters, mobile phones, scissors, light bulbs, cans, glasses, balls, markers, cups, and remote controls.
This dataset has been gathered in 11 separate sessions, with 8 of them being indoor sessions and 3 being outdoor ones. These sessions feature diverse backgrounds and lighting conditions.  
The Core50 dataset consists of 164,866 128×128 RGB images, which is computed based on the formula of 11 sessions×50 objects×300 frames per session. To align with the settings of the ImageNet-100 dataset, we resize the images in the Core50 dataset to 224×224 during data preprocessing.
In this paper, we split 50 classes into 10 tasks to conduct class incremental learning, each task contains 5 classes. We select session 11 for testing and the remaining 10 sessions are used for training. Since each class in the Core50 dataset contains only one object, the classification task difficulty on this dataset is lower compared to ImageNet-100. We set the number of saved samples to 10 per class and 500 in total for other replay-based methods, thus the extra memory costs were measured as 72 MB for Replay and iCarl, and 457 MB for DER. In BrainCL, the number of memory cues transferred into long-term memory is set to 100 per class, and the extra memory cost is 70 MB.

\vspace{5pt}
\noindent\textbf{Medical Dataset.} 
MedMINIST\cite{yang2023medmnist} is a large-scale MNIST-like dataset collection of standardized biomedical images. It contains 12 datasets for 2D and 6 datasets for 3D, consisting of 708,069 2D images and 9,998 3D images in total. All of the images are released in 4 different resolutions: 28×28, 64×64, 128×128 and 224×224. We select 7 2D datasets from MedMINIST to develop a medical dataset containing 40 classes in total, the resolutions of selected datasets are 224×224.
However, the quantities of training images in different datasets and different classes vary widely (less than 100 to over 10,000), which could cause severe class imbalance. To tackle this problem, we decide to limit the maximum quantity of training images in single class to 1,000: for those classes containing more than 1,000 training images, we randomly keep 1,000 images and eliminate others. Moreover, we keep at most 50 test images for each class for efficient evaluation. Thus the total quantity of the training samples is 34,554 and the quantity of test samples is 1,944. For class incremental learning scenario, we split 40 classes into 10 tasks and each task contains 4 classes. Similar to the setting in ImageNet-100, we set the number of saved samples to 20 per class and 800 in total, the extra memory cost is 115 MB for Replay and iCarl, 500 MB for DER. In BrainCL, the number of memory cues transferred into long-term memory is set to 1000 per class, and the extra memory cost is 97 MB.

\section*{Data and code availability}

\subsection*{Computational Resources}

\section*{Acknowledgments}

\section*{Author contributions}

\section*{Competing interests}

The authors have no competing interests.

\bibliography{references}

\renewcommand{\thefigure}{S\arabic{figure}}
\setcounter{figure}{0}


\end{document}